\documentclass[a4paper,fleqn]{cas-dc}

\usepackage[numbers]{natbib}
\usepackage{capt-of}
\usepackage{placeins}
\def\tsc#1{\csdef{#1}{\textsc{\lowercase{#1}}\xspace}}
\tsc{WGM}
\tsc{QE}
\tsc{EP}
\tsc{PMS}
\tsc{BEC}
\tsc{DE}
\begin{document}
% Begin inlined from generated/result_macros.tex
\newcommand{\AgentTaskPass}{57.9\%}
\newcommand{\AgentAvgCheck}{74.6\%}
\newcommand{\AgentGlobalCheck}{72.0\%}
\newcommand{\AgentTotalFailed}{12.9\%}
\newcommand{\ReactTaskPass}{30.6\%}
\newcommand{\ReactAvgCheck}{47.9\%}
\newcommand{\ReactGlobalCheck}{43.1\%}
\newcommand{\ReactTotalFailed}{32.4\%}
% End inlined from generated/result_macros.tex
\let\WriteBookmarks\relax
\shorttitle{MultiUAV-Plat}
\shortauthors{Zhang et~al.}

\title [mode = title]{MultiUAV-Plat: An LLM-Oriented Platform, Benchmark and Framework for Multi-UAV Collaborative Task Planning}                      
% \tnotemark[1,2]

% \tnotetext[1]{This document is the results of the research
%    project funded by the National Science Foundation.}

% \tnotetext[2]{The second title footnote which is a longer text matter
%    to fill through the whole text width and overflow into
%    another line in the footnotes area of the first page.}

\author[1]{Sheng Zhang}
% \cormark[1]
\ead{zhangsheng@nudt.edu.cn}
\credit{Conceptualization of this study, Methodology, Software}

% \address[1]{, Street 129, 1043 NX Amsterdam, The Netherlands}
\affiliation[1]{organization={National Key Laboratory of Information Systems Engineering, National University of Defense Technology},
                addressline={Deya Road}, 
                city={Changsha},
                postcode={410073}, 
                state={Hunan},
                country={China}}

\author[1]{Qinglin Li}
\ead{liqinglin@nudt.edu.cn}
\author[1]{Yuechao Zang}
\ead{zangyuechao@nudt.edu.cn}
\author[1]{Xueqin Huang}
\ead{hxq@nudt.edu.cn}
\author[1]{Yijia Fu}
\ead{fuyijia20@nudt.edu.cn}

\author[1]{Cheng Zhu}
\cormark[1]
\ead{zhucheng@nudt.edu.cn}
\cortext[cor1]{Corresponding author}
% \fntext[fn1]{This is the first author footnote.}
% \fntext[fn2]{Another author footnote.}

\nonumnote{The platform is open-sourced at \url{https://github.com/zhangsheng93/MultiUAV-Plat}}

\begin{abstract}
Large language models (LLMs) provide a promising interface for high-level robotic task planning, but their use in multi-UAV collaboration remains difficult to evaluate systematically. Existing UAV simulators mainly emphasize dynamics, perception, or low-level control, while existing LLM-agent benchmarks rarely capture aerial-robotics constraints such as partial observability, spatial coverage, UAV assignment, and multi-vehicle coordination. To bridge this gap, we present \textit{MultiUAV-Plat}, a lightweight, easy-to-use, LLM-agent-oriented simulation platform for multi-UAV collaborative task planning. The platform exposes concise RESTful APIs, agent-facing observations, role-based information access, hidden validation logic, and optional 2D/3D visualization, allowing agents to solve missions through realistic tool interaction rather than privileged simulator access. Built on this platform, the \textit{MultiUAV-Plat Benchmark} contains 75 mission sessions, 1500 natural-language tasks, and 9396 validation checks across target assignment, area search, and area assignment and patrol scenarios. We further propose \textit{Agent4Drone}, a task-specific LLM agent framework that structures multi-UAV behavior into memory, observation, task understanding, planning, execution, and verification. In a full paired benchmark comparison, Agent4Drone achieves a \AgentTaskPass{} task pass rate, a \AgentAvgCheck{} average task check pass rate, and a \AgentGlobalCheck{} global check pass rate, substantially outperforming a ReAct baseline at \ReactTaskPass{}, \ReactAvgCheck{}, and \ReactGlobalCheck{}, respectively. Agent4Drone also reduces the total failed task rate from \ReactTotalFailed{} to \AgentTotalFailed{}. These results demonstrate that MultiUAV-Plat and MultiUAV-Plat Benchmark provide a reproducible foundation for studying LLM-driven multi-UAV autonomy under realistic information and execution constraints.

\end{abstract}

% \includegraphics{figs/cas-grabs.pdf}

% \begin{highlights}
% \item Research highlights item 1
% \item Research highlights item 2
% \item Research highlights item 3
% \end{highlights}


\maketitle
\hypersetup{
  pdftitle={MultiUAV-Plat: An LLM-Oriented Platform, Benchmark and Framework for Multi-UAV Collaborative Task Planning},
  pdfauthor={Sheng Zhang, Qinglin Li, Yuechao Zang, Xueqin Huang, Yijia Fu, Cheng Zhu}
}

\section{Introduction}

Multiple Unmanned Aerial Vehicle (multiUAV) systems have emerged as a transformative technology across a wide spectrum of real-world applications, including environmental monitoring, aerial searching, emergency response, and logistics. Their ability to provide rapid coverage, flexible deployment, and scalable sensing capabilities has made them indispensable in both civilian and industrial contexts. Despite this growing prevalence, the control and coordination of multi-UAV systems remain highly challenging. Existing mission-planning and path-planning studies have developed strong foundations for assignment, routing, coverage, obstacle avoidance, and energy-aware coordination~\cite{song2023mission,rahman2025multiuavsurvey,kong2024multiuav}. However, traditional approaches typically rely on expert-engineered problem formulations and labor-intensive manual control pipelines, which often result in suboptimal efficiency, limited adaptability, and high operational overhead. As mission complexity increases, these limitations become even more pronounced.

The advent of large language models (LLMs) offers a promising paradigm shift for intelligent multi-UAV coordination. Equipped with strong natural language understanding, extensive world knowledge, and increasingly capable tool-use interfaces, LLM-based agents have demonstrated significant potential to reason about task structure, plan high-level strategies, and execute complex action sequences~\cite{yao2023react,ahn2022saycan,liang2022code}. Recent multi-robot and multi-UAV studies further suggest that LLMs can decompose high-level instructions, form coalitions, coordinate multiple robots, and support natural-language aerial control~\cite{kannan2023smartllm,mandi2023roco,liu2024promptdriven,nazzari2026tacos}. However, it remains unclear whether these abilities transfer to multi-UAV missions that require closed-loop interaction with restricted robotic interfaces. A realistic LLM-controlled UAV agent cannot assume full simulator state: it must interpret natural-language objectives, choose suitable UAVs, actively collect missing local observations, issue executable API calls, coordinate multiple vehicles, and verify progress under hidden mission conditions. This leads to the central research question of this paper: \textit{How can LLM agents be systematically evaluated for multi-UAV collaborative task planning when they operate through restricted UAV APIs, partial local perception, and hidden task validators?}

\begin{figure*}[htbp]
\centering
\begin{tabular}{cc}
\includegraphics[height=0.22\textheight]{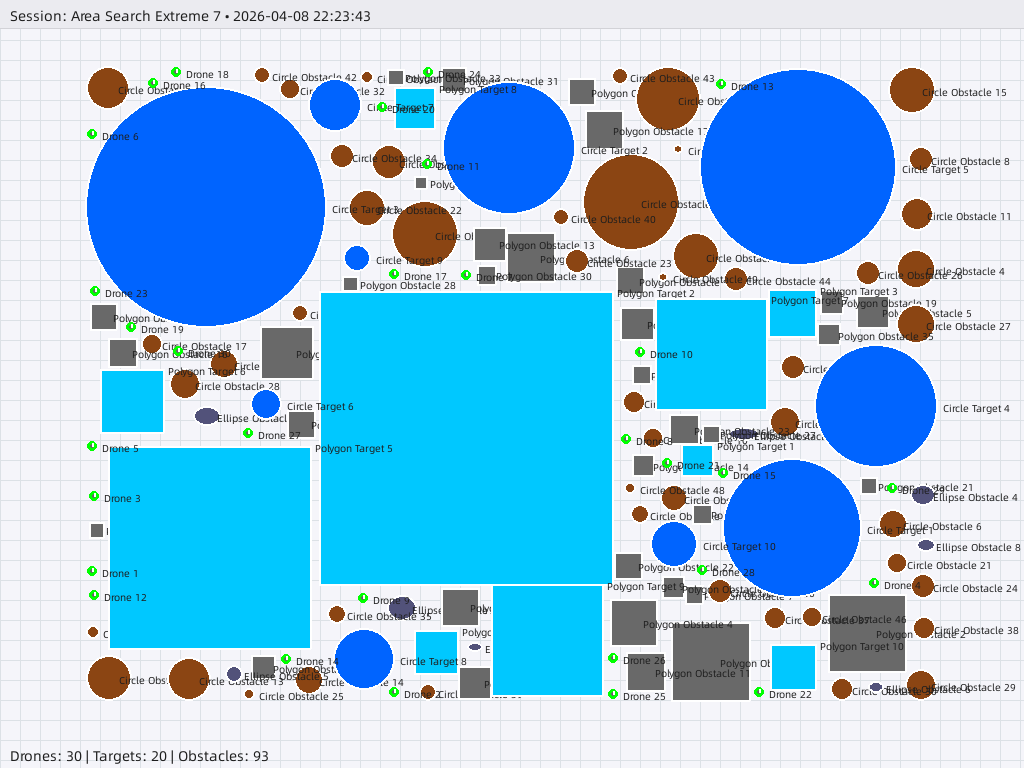} &
\includegraphics[height=0.22\textheight]{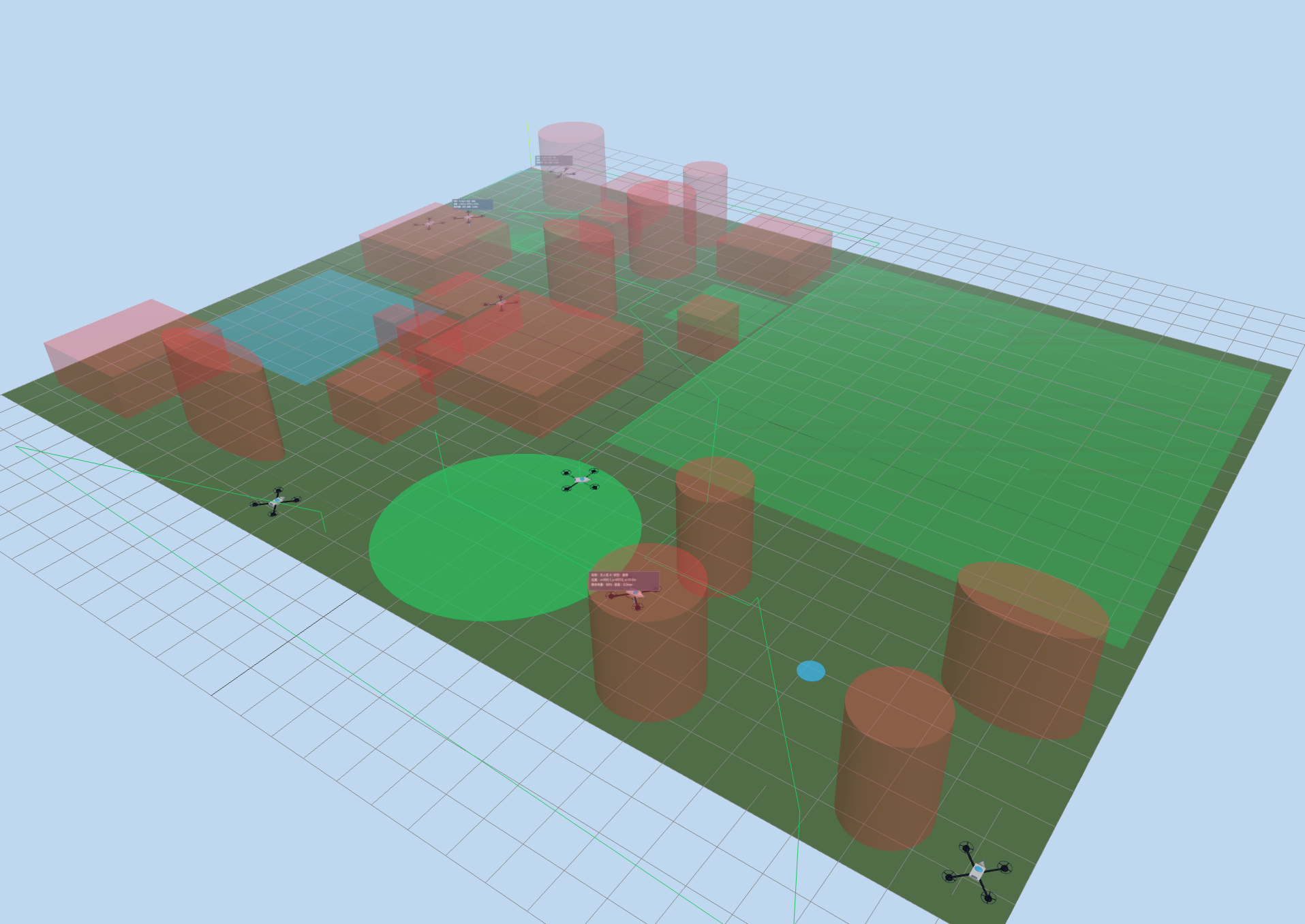} \\
\textbf{2D overview view} & \textbf{3D visualization view}
\end{tabular}
\caption{MultiUAV-Plat platform views for UAV task planning. The 2D and 3D views represent the same underlying mission state. The 2D view is designed for quick overview, scenario inspection, and task-level spatial reasoning, while the 3D view is designed for intuitive visualization of UAV execution and environment geometry.}
\label{fig:platform_views}
\end{figure*}

To address this gap, we introduce MultiUAV-Plat, a lightweight, easy-to-use, open, and extensible simulation platform specifically designed for LLM-agent-driven multi-UAV collaborative task planning and control. Figure~\ref{fig:platform_views} shows the platform's synchronized 2D overview and 3D visualization views for the same mission state. Unlike high-fidelity UAV simulators that mainly expose physics, perception, and control interfaces~\cite{shah2017airsim,song2020flightmare,panerati2021learning}, the platform provides a unified, LLM-friendly API interface through which agents can issue UAV actions, receive agent-facing observations, and iteratively adjust their plans according to changing mission states without interacting with simulator internals or low-level flight-control code. By abstracting low-level system integration while preserving closed-loop action control, MultiUAV-Plat enables researchers to evaluate not only high-level planning, but also the execution, feedback handling, and replanning capabilities of intelligent agents. To the best of our knowledge, MultiUAV-Plat is the first LLM-agent-oriented simulation and benchmarking platform for executable multi-UAV collaborative task planning. It evaluates agents through restricted UAV APIs, partial local observations, closed-loop execution, and hidden mission-level validators. The benchmark covers a wide range of difficulty levels, from simple API invocation tests to challenging single-UAV tasks and fully collaborative multi-UAV missions.

Beyond the platform and benchmark, we further propose Agent4Drone, a framework specialized for multi-UAV collaborative task planning. Designed to enhance LLM generalization across diverse mission profiles, Agent4Drone provides structured guidance that helps LLM agents decompose tasks, manage tool calls, and coordinate multiple UAVs more effectively. Experimental results demonstrate that Agent4Drone improves task pass rate by 27.33 percentage points over ReAct on the full benchmark. This establishes a strong task-specific framework and provides a practical starting point for future research in LLM-controlled aerial robotics.

In summary, this work contributes: (1) MultiUAV-Plat, a lightweight, easy-to-use, LLM-agent-oriented simulation platform for multi-UAV collaborative task planning; (2) MultiUAV-Plat Benchmark, a benchmark suite for reproducible evaluation of LLM agents under restricted API access, partial local perception, and hidden validation; and (3) Agent4Drone, a reference LLM-agent framework with a full paired benchmark comparison against ReAct. Together, these contributions lay the groundwork for an emerging research direction at the intersection of large-scale language models and autonomous multi-robot systems.

% The contributions of the papaer are:
% \begin{itemize} 
% \item keywords and MSC codes 
% \item theorems, definitions and
% \item lables of enumerations 
% \item citation style and labeling.
% \end{itemize}

% This class depends on the following packages
% for its proper functioning:

% \begin{enumerate}
% \itemsep=0pt
% \item {natbib.sty} for citation processing;
% \item {geometry.sty} for margin settings;
% \item {fleqn.clo} for left aligned equations;
% \item {graphicx.sty} for graphics inclusion;
% \item {hyperref.sty} optional packages if hyperlinking is
%   required in the document;
% \end{enumerate}  

\section{Related Work}

\subsection{Multi-UAV Task Planning}

Research on multi-UAV systems has long studied task planning as the coupled problem of task assignment, path planning, coverage, coordination, and constraint satisfaction. Recent surveys on multi-UAV mission planning~\cite{song2023mission} and multi-UAV path planning~\cite{rahman2025multiuavsurvey} summarize a broad family of methods, including mathematical programming, heuristics and metaheuristics, negotiation-based allocation, and learning-based approaches. These methods provide important foundations for target assignment, area coverage, collision avoidance, energy-aware routing, and coordinated execution. Kong et~al.~\cite{kong2024multiuav} further formulate simultaneous target assignment and path planning under dynamic obstacles and partial observability as a learning problem, using deep reinforcement learning to combine assignment and motion decisions. However, most existing studies assume a predefined optimization objective, a specialized controller, or direct access to structured state information. This leaves open the problem of evaluating whether a language-based agent can interpret natural-language tasks, actively gather missing information, use restricted platform APIs, and coordinate multiple UAVs through closed-loop tool interaction.

\subsection{UAV and Swarm Simulation Platforms}

Simulation platforms are another important prerequisite for UAV research. General robotic and UAV simulators such as AirSim~\cite{shah2017airsim}, Flightmare~\cite{song2020flightmare}, and gym-pybullet-drones~\cite{panerati2021learning} provide high-fidelity visual simulation, fast quadrotor dynamics, reinforcement-learning interfaces, and multi-agent control support. Chen et~al.~\cite{chen2023uavswarm} further compare open-source UAV swarm simulation environments such as Webots, Gazebo, CoppeliaSim, ARGoS, MORSE, and related tools with respect to realism, sensors, performance, programming interfaces, and swarm support. These platforms have significantly advanced algorithm development and sim-to-real experimentation, but their primary abstractions are still physics, perception, control, or reinforcement-learning tasks. In contrast, LLM-oriented multi-UAV research requires a different interface layer: natural-language task instructions, agent-visible state boundaries, simple tool-call APIs, hidden validation logic, and reproducible benchmark sessions that evaluate mission-level reasoning rather than only low-level control.

\begin{table*}[htbp]
\centering
\caption{Comparison with representative simulators, benchmarks, and frameworks. ``Hidden task validation'' denotes automatic platform-side validation of executable mission outcomes rather than offline question answering, conversational compliance, or low-level reward only.}
\label{tab:benchmark_comparison}
\small
\setlength{\tabcolsep}{3pt}
\resizebox{\textwidth}{!}{%
\begin{tabular}{lccccccc}
\hline
\textbf{Platform / Benchmark / Framework} & \textbf{Domain} & \textbf{LLM-facing} & \textbf{Multi-agent} & \textbf{UAV / Aerial} & \textbf{Partial obs.} & \textbf{Tool/API actions} & \textbf{Hidden task validation} \\
\hline
AirSim~\cite{shah2017airsim} & UAV / AV simulation & No & Partial & Yes & No & Yes & No \\
Flightmare~\cite{song2020flightmare} & Quadrotor simulation & No & Partial & Yes & No & Yes & No \\
gym-pybullet-drones~\cite{panerati2021learning} & Multi-drone RL & No & Yes & Yes & No & Yes & Reward-based \\
ALFWorld~\cite{shridhar2021alfworld} & Household embodied AI & Yes & No & No & Partial & Yes & Yes \\
WebShop~\cite{yao2022webshop} & Web interaction & Yes & No & No & Partial & Yes & Yes \\
AgentBench~\cite{liu2023agentbench} & General LLM agents & Yes & Partial & No & Partial & Yes & Mixed \\
EmbodiedBench~\cite{du2025embodiedbench} & Multimodal embodied AI & Yes & No & No & Yes & Yes & Yes \\
SafeAgentBench~\cite{yin2024safeagentbench} & Safe embodied planning & Yes & Yes & No & Yes & Yes & Yes \\
MAP-THOR~\cite{nayak2024mapthor} & Multi-agent household planning & Yes & Yes & No & Yes & Yes & Yes \\
SMART-LLM~\cite{kannan2023smartllm} & Multi-robot task planning & Yes & Yes & No & No & No & Dataset \\
RoCo~\cite{mandi2023roco} & Multi-robot collaboration & Yes & Yes & No & Partial & Yes & Task-based \\
AirCopBench~\cite{zha2026aircopbench} & Multi-drone perception & Yes & Yes & Yes & Yes & No & QA-based \\
MM-UAVBench~\cite{dai2025mmuavbench} & UAV multimodal reasoning & Yes & No & Yes & Yes & No & QA-based \\
UAVBench~\cite{ferrag2025uavbench} & UAV scenario dataset & Yes & Partial & Yes & No & No & Risk labels \\
$\alpha^3$-Bench~\cite{ferrag2026alpha3bench} & UAV networked agents & Yes & Yes & Yes & Partial & Yes & Conversational \\
Prompt-driven multi-drone planning~\cite{liu2024promptdriven} & Multi-drone task planning & Yes & Yes & Yes & Partial & No & No \\
UAV-CodeAgents~\cite{sautenkov2025uavcodeagents} & UAV mission generation & Yes & Yes & Yes & Partial & Partial & No \\
TACOS~\cite{nazzari2026tacos} & Multi-drone natural-language control & Yes & Yes & Yes & No & Yes & Demonstration \\
\textit{Web-of-Drones}~\cite{iannoli2026webofdrones} & UAV swarm execution & Yes & Yes & Yes & No & Yes & No \\
\textbf{MultiUAV-Plat} & \textbf{Multi-UAV task planning} & \textbf{Yes} & \textbf{Yes} & \textbf{Yes} & \textbf{Yes} & \textbf{Yes} & \textbf{Yes} \\
\hline
\end{tabular}
}
\end{table*}

\subsection{LLM Agents and Embodied Benchmarks}

Large language models have recently been evaluated as interactive agents that combine reasoning, tool use, and environment feedback. Benchmarks such as ALFWorld~\cite{shridhar2021alfworld} and WebShop~\cite{yao2022webshop} evaluate grounded language agents in interactive household or web environments, and AgentBench~\cite{liu2023agentbench} broadens this direction by measuring LLM agents across multiple decision-making settings. Recent embodied-agent benchmarks further evaluate multimodal embodied reasoning, safety-aware task planning, and multi-agent planning under partial observability, including EmbodiedBench~\cite{du2025embodiedbench}, SafeAgentBench~\cite{yin2024safeagentbench}, and MAP-THOR~\cite{nayak2024mapthor}. Classical pathfinding resources such as the MovingAI benchmarks~\cite{sturtevant2012benchmarks} and MAPF benchmarks~\cite{stern2019mapf} provide widely used grid-based and multi-agent path-finding maps and scenarios for algorithm comparison, but they focus on pathfinding instances rather than LLM-facing, partially observable, executable UAV tasks.

Recent aerial-agent benchmarks are closer to multi-UAV settings. AirCopBench~\cite{zha2026aircopbench} evaluates multi-drone collaborative embodied perception and reasoning for multimodal large language models. MM-UAVBench~\cite{dai2025mmuavbench} evaluates multimodal perception, cognition, and planning over real low-altitude UAV data. UAVBench~\cite{ferrag2025uavbench} provides structured UAV scenarios and multiple-choice reasoning tasks, while $\alpha^3$-Bench~\cite{ferrag2026alpha3bench} evaluates conversational UAV agents with tool calls and agent-to-agent coordination under network, safety, robustness, and efficiency constraints. Compared with these benchmarks, MultiUAV-Plat Benchmark focuses on closed-loop executable multi-UAV task planning: agents interact through restricted APIs, receive partial observations, and are evaluated by hidden task validators. Table~\ref{tab:benchmark_comparison} summarizes this difference among representative simulators, agent benchmarks, embodied benchmarks, aerial benchmarks, and multi-robot or multi-drone frameworks.

\subsection{LLM-Based Multi-Robot and Aerial Agents}

Basic LLM-agent frameworks provide general mechanisms for reasoning and action selection. Chain-of-thought prompting~\cite{wei2022cot} elicits intermediate reasoning steps for complex tasks, while Tree of Thoughts~\cite{yao2023tot} and Graph of Thoughts~\cite{besta2023got} extend this idea by searching over branching or graph-structured reasoning states. ReAct~\cite{yao2023react} interleaves reasoning traces with actions and demonstrates that language models can improve interactive decision making by repeatedly observing and acting in an environment. These frameworks are useful general foundations, but they do not by themselves specify how to assign UAVs, gather local aerial observations, coordinate multiple vehicles, or verify executable mission progress.

Robotics-oriented LLM frameworks add grounding between language and action. SayCan~\cite{ahn2022saycan} grounds language-model planning in affordance scores, Code as Policies~\cite{liang2022code} uses code-generation models to synthesize robot policy programs from natural-language instructions, and Sun et~al.~\cite{sun2024interactive} study LLM-based task planning under partial observability, where the robot must collect missing information before choosing actions. LLM-based multi-robot studies further use language models for decomposition, role assignment, communication, and coordination. SMART-LLM~\cite{kannan2023smartllm} decomposes high-level instructions into multi-robot task plans through task decomposition, coalition formation, and task allocation. RoCo~\cite{mandi2023roco} uses LLM-mediated dialogue for multi-robot collaboration and includes environmental feedback such as collision checking for plan refinement. Recent surveys and frameworks further study LLMs for multi-robot task allocation, dependency modeling, heterogeneous collaboration, and end-to-end execution, including the multi-robot LLM survey by Li et~al.~\cite{li2025llmmrs}, LiP-LLM~\cite{obata2024lipllm}, COHERENT~\cite{liu2024coherent}, and DART-LLM~\cite{wang2024dartllm}. In the aerial domain, prompt-driven multi-drone planning~\cite{liu2024promptdriven} explores natural-language task planning, UAV-CodeAgents~\cite{sautenkov2025uavcodeagents} combines multi-agent ReAct with vision-language grounding for UAV mission generation, and TACOS~\cite{nazzari2026tacos} executes multi-drone plans through a library of UAV APIs. More recently, the Web-of-Drones framework~\cite{iannoli2026webofdrones} has demonstrated function-calling-based closed-loop UAV swarm execution through standardized MCP and Web-of-Things interfaces. These systems establish that LLMs can plan and execute selected multi-UAV missions, but they are evaluated as task-specific frameworks rather than reusable benchmark suites with large task collections and hidden mission-level validators. This distinction motivates an executable benchmark for systematic comparison under consistent task, observation, action, and grading protocols.

\section{Methodology}

\begin{figure*}[htbp]
\centering
% \small
% \setlength{\fboxsep}{6pt}
% \renewcommand{\arraystretch}{1.15}
% \begin{tabular}{ccc}
% \fbox{\begin{minipage}{0.27\textwidth}
% \centering
% \textbf{Agent4Drone}\\[2pt]
% Observe $\rightarrow$ Understand $\rightarrow$ Plan\\
% $\rightarrow$ Execute $\rightarrow$ Verify\\[2pt]
% \footnotesize
% Agent-access context, UAV selection, short-horizon planning, bounded replanning
% \end{minipage}}
% &
% \raisebox{1.5em}{$\Longleftrightarrow$}
% &
% \fbox{\begin{minipage}{0.27\textwidth}
% \centering
% \textbf{Benchmark}\\[2pt]
% 75 sessions, 1500 tasks, 9396 checks\\
% 3 scenario families, 5 difficulty levels\\[2pt]
% \footnotesize
% Partial observation and hidden validation for reproducible evaluation
% \end{minipage}}
% \\[1.2em]
% \multicolumn{3}{c}{$\Updownarrow$}\\[-0.2em]
% \multicolumn{3}{c}{
% \fbox{\begin{minipage}{0.78\textwidth}
% \centering
% \textbf{MultiUAV-Plat}\\[2pt]
% \begin{tabular}{c}
% \textbf{Service Layer}: LLM-friendly interaction interface and session context\\
% \textbf{Control Layer}: command execution, state update, conflict and safety logic\\
% \textbf{Model Layer}: Drone, Target, Obstacle, Environment, Session, and Task\\
% \textbf{Security / Permission}: agent-level visibility, hidden checks, restricted global access
% \end{tabular}
% \end{minipage}}
% }
% \end{tabular}
\includegraphics[width=0.9\textwidth]{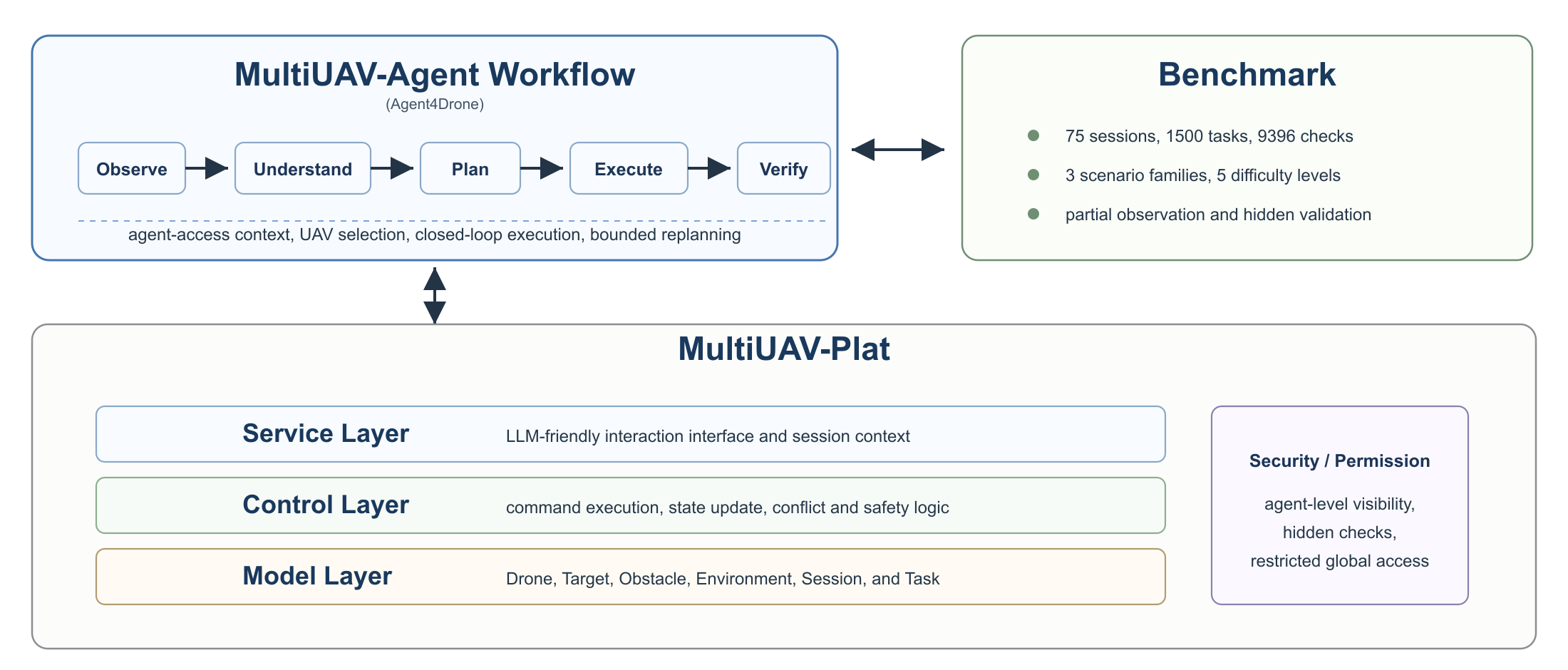}
\caption{Methodology overview. MultiUAV-Plat provides the interaction substrate, the MultiUAV-Plat Benchmark supplies standardized tasks and automatic validation, and the proposed MultiUAV-Agent Workflow, named Agent4Drone, performs closed-loop multi-UAV task planning under agent-level partial observability.}
\label{fig:methodology_overview}
\end{figure*}

As illustrated in Fig.~\ref{fig:methodology_overview}, our methodology consists of three closely connected components: platform, benchmark, and workflow framework. The platform provides the simulation substrate, the benchmark defines the evaluation protocol, and Agent4Drone instantiates the proposed MultiUAV-Agent Workflow for LLM-agent research on multi-UAV collaboration.

\subsection{Platform}

To systematically address the challenges in the development and evaluation of multi-UAV collaborative task planning and control driven by large language models, we propose \textit{MultiUAV-Plat}, a lightweight, modular, open-source, and LLM-agent-oriented simulated interactive environment for multi-UAV systems. The platform is designed for low setup cost and direct agent integration: agents receive natural-language task instructions, agent-facing observations, and environment feedback; issue concrete UAV actions through concise RESTful APIs and simple action primitives; observe state changes; and adjust subsequent plans according to the evolving mission context. The core architecture of the platform consists of three main layers, a cross-cutting security and permission management mechanism, and an optional visual simulation interface. Together, these components establish a complete loop from physical entities to intelligent control, providing an easy-to-use experimental environment for developing, testing, and evaluating LLM-based autonomous agents without requiring users to work with simulator internals or low-level flight-control code.

The Model Layer, as the foundation of the platform, provides formalized definitions and encapsulation of core entities in the system and serves as the minimal interactive unit of the simulation for the upper layer. This layer primarily consists of six models: Drone, Target, Obstacle, Environment, Session, and Task. The first four are concrete models, while the latter two are abstract models.

The drone model is the primary interactive entity for LLM agents, encapsulating the drone's dynamic properties, real-time status information such as position, velocity, battery level, and sensor states, as well as basic action commands including takeoff, landing, and waypoint navigation. The target model provides a generalized representation of task objectives, which can be either static or dynamic discrete target points or target areas requiring inspection or coverage. This flexible model design effectively supports diverse task scenarios, from simple navigation to complex area patrol. The obstacle model defines static threats in the environment, such as buildings, mountains, and no-fly zones. It includes geometric shapes and spatial occupancy information, serving as constraints that must be avoided in UAV path planning or collision detection. The Environment model represents the macro-level physical conditions, such as wind, weather, and illumination, thereby enhancing the realism and uncertainty of the simulation.  The session model instantiates a complete task scenario, aggregating all the aforementioned model entities and defining their spatial and logical relationships. The session model contains a series of task models, providing specific, natural language task instructions for pre-configured scenarios within the session, thereby offering clear planning and execution objectives for LLM-based agents.

Above the model layer, the control layer serves as the logical processing core of the platform. It is responsible for the execution of low-level operations, model state updates, and decision arbitration to ensure coherent simulation and effective task execution.  Its functions primarily include lifecycle management of model entities, providing standardized operational management such as creation, update, query, and deletion for all model entities. It also handles complex logical judgments and processing, such as collision detection between UAVs and obstacles or among UAVs themselves, logical judgments of conflicts in task allocation among UAVs, as well as continuous states tracking of subtask execution.

The service layer provides standardized application programming interfaces (APIs) that connect the platform to external intelligent agents. Designed with an LLM-friendly philosophy, it offers RESTful-compliant APIs that enable agents to control UAVs through concise, tool-like calls rather than simulator internals or low-level flight-control code. These APIs expose action primitives such as observation, takeoff, landing, waypoint navigation, sensing, photography, and verification feedback in a form that can be directly used by LLM agents. Additionally, for each active agent, the service layer maintains an independent session context that supports multi-turn complex task planning and decision-making.

The security and permission management mechanism controls platform access through request-level authorization and role-based information boundaries. Each API request is authenticated and checked against its permission level, so that privileged users may access scenario configuration and global management functions, while LLM agents are restricted to agent-level operation interfaces. In particular, agent requests return only information within the local perception range of the corresponding UAVs, rather than exposing global targets, obstacles, or hidden validation data. This permission design improves system safety and reliability, prevents agents from using privileged global information, and better reflects real-world UAV operation, where each UAV must make decisions based on limited local perception.

The optional user interface further enhances the usability of the platform by providing synchronized 2D overview and 3D visualization views. The 2D interface, implemented with pygame, dynamically displays the real-time positions and statuses of UAVs, targets, and obstacles as well as interactions between external agents and simulated entities. The 3D visualization complements this top-down view by rendering UAV execution, target and obstacle geometry, and spatial relationships in an intuitive three-dimensional scene. Together, these views offer users direct feedback on task execution progress and facilitate scenario inspection, algorithm debugging, performance analysis, and result verification for efficient human-in-the-loop LLM agent framework development.

Overall, MultiUAV-Plat provides a set of distinctive advantages for research on LLM-driven multi-UAV collaboration. Its lightweight implementation, preconfigured sessions, reusable templates, API documentation, optional 2D/3D visualization, and hidden validators make it easy to build, test, and compare LLM agents. Its LLM-friendly API design enables researchers to focus on decision-making algorithms without engaging with low-level simulation details, while the visualization layer makes both abstract mission states and three-dimensional execution processes easier to inspect. Its human--machine interaction capabilities support both fully autonomous UAV control and manual intervention for scenario construction, task adjustment, and real-time monitoring. The platform also offers reusable scenario resources and a task-template mechanism that guide users in configuring scenarios, designing tasks, and rapidly generating customized test cases. In addition, its robust permission system ensures data isolation and operational safety in multi-user and multi-agent scenarios, making the platform well-suited for collaborative research and large-scale experiments.

\subsection{Benchmark}

To complement the platform, we construct the \textit{MultiUAV-Plat Benchmark}, a benchmark suite for evaluating LLM agents in multi-UAV collaborative task planning. To the best of our knowledge, MultiUAV-Plat Benchmark is the first executable benchmark specifically designed to evaluate LLM agents on multi-UAV collaborative task-planning missions under partial observability. Unlike offline question-answering or conversational evaluation, agents must act through restricted UAV APIs and satisfy hidden task-level validators. The benchmark contains 75 mission sessions, 1500 natural-language tasks, and 9396 hidden validation checks. In contrast to benchmarks that evaluate either isolated language understanding or low-level motion control, the benchmark focuses on the mission-planning layer where natural-language reasoning, tool use, spatial decision-making, and multi-agent coordination must be integrated.

The benchmark is designed to address the gap between language-agent evaluation and realistic multi-UAV mission execution. In practical UAV operations, an agent must not only understand a natural-language objective, but also select appropriate UAVs, issue executable actions, interpret local perception feedback, and revise its plan under incomplete information. These requirements are difficult to measure with static language tasks or full-information simulators. Therefore, the benchmark emphasizes three design goals: evaluating mission-level reasoning through natural-language task instructions, enforcing agent-level partial observability through restricted platform interfaces, and supporting reproducible scoring through hidden automatic validation. This design allows the benchmark to measure whether an LLM agent can complete UAV missions through closed-loop interaction rather than relying on privileged global state or purely textual reasoning.

\subsubsection{Benchmark Design}

Each benchmark session defines a complete mission scenario, including UAVs, targets, obstacles, environment settings, and a set of task instructions. Each task is scored as an individual unit, while tasks within a session are executed sequentially. Before the next task, the battery state is reset and all UAVs are forced to land, but their final positions are retained. Consequently, a preceding task may affect the initial UAV positions of the following task, although the pass or failure of each task is determined separately by its own validation conditions. The final benchmark score is computed from these task-level outcomes rather than from a required task order.

The benchmark follows an agent-facing protocol rather than an omniscient simulator protocol. An LLM agent can observe task descriptions, inspect UAV states, issue UAV actions, and receive only local perception information around each UAV. It cannot directly perceive global targets or obstacles; instead, targets and obstacles become visible only when they fall within a UAV's perceived radius. This partial-observation setting makes the benchmark more realistic and substantially more challenging than full-information planning: agents must actively gather information, reason from incomplete local observations, and coordinate UAVs without directly reading the full mission map.

This design also separates task solving from task grading. Tasks are presented to the agent as natural-language objectives, while task completion is evaluated by platform-side automatic checks. They are hidden validation points used only for reproducible scoring after task planning and execution. The benchmark contains 9396 hidden checks across 1500 tasks: each task has 1--39 checks, with a median of 4 and a mean of 6.26. As shown in Fig.~\ref{fig:check_count_distribution}, tasks with more checks contain more conditions that must be satisfied and therefore generally require more execution, observation, and verification steps. The check count is not an exact measure of trajectory length, because several conditions may be satisfied by one action and one condition may require multiple actions. Across 21 endpoint types, the checks cover movement, perception, target, area, state, and coordination requirements, making success dependent on sustained state tracking, multi-step execution, and avoiding partial completion failures.

\begin{center}
\centering
\includegraphics[width=\columnwidth]{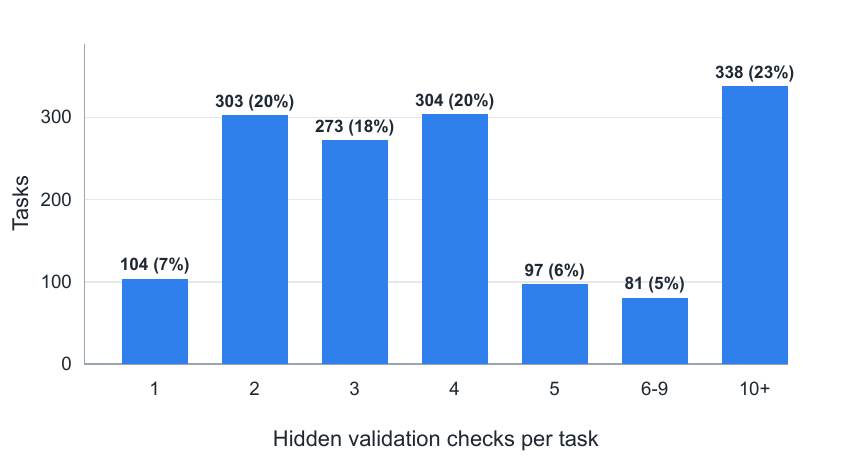}
\captionof{figure}{Distribution of hidden validation-check counts per task. More checks generally correspond to more task conditions and more execution, observation, and verification steps, although check count is not an exact trajectory-length measure.}
\label{fig:check_count_distribution}
\end{center}

\subsubsection{Scenario Families and Difficulty Design}

The benchmark contains three scenario families: \textit{Area Search}, \textit{Area Assignment and Patrol}, and \textit{Target Assignment}. Each family contains 25 sessions and 500 tasks. Within each family, sessions are divided into five difficulty levels: Easy, Intermediate, Moderate, Hard, and Extreme. Each difficulty level contains five independently seeded sessions, and each session contains 20 tasks.

The scenario family indicates the dominant mission configuration and evaluation emphasis of a session, but it does not restrict the session to containing only one type of task. For example, an area-search session may still include basic navigation, photo-taking, landing, or target-reaching subtasks. This mixed composition better reflects realistic multi-UAV missions, where high-level objectives are usually achieved through heterogeneous operational subtasks.

\textit{Area Search} evaluates spatial exploration and coverage over extended area targets. Agents must direct one or more UAVs to fully search circular or polygonal regions, inspect the relevant area geometry, and complete coverage-oriented objectives without assuming full prior knowledge of the environment. This family tests whether LLM agents can move beyond point-to-point navigation toward systematic information gathering and region-level search behavior.

\textit{Area Assignment and Patrol} combines point-target assignment with area-search objectives. Agents must coordinate UAVs across mixed mission targets, including discrete points that must be reached or searched and circular or polygonal regions that require coverage or patrol. This family emphasizes division of labor, multi-UAV coordination, and maintaining structured behavior when heterogeneous target types must be handled within the same shared environment.

\textit{Target Assignment} evaluates whether agents can allocate many targets across multiple UAVs and complete target-oriented objectives such as reaching, inspecting, photographing, delivering to, or returning from those targets. This family emphasizes assignment and execution over discrete spatial objectives, especially when many UAVs and candidate targets are present but only local perception is accessible.

The benchmark difficulty is defined at the session level and reflects the overall complexity of the mission world. As difficulty increases, sessions generally involve larger fleets, more targets, denser obstacles, and larger mission areas. Across all sessions, the number of UAVs ranges from 3 to 30, the number of targets ranges from 3 to 100, and the number of obstacles ranges from 6 to 100. In addition, the target and obstacle geometry varies across sessions: targets may differ in shape, size, and spatial extent, while obstacles differ in shape, size, density, and distribution. These variations force agents to adapt their planning strategies to different spatial layouts rather than relying on a fixed mission template.

Individual tasks also carry local task-complexity labels, such as easy, medium, and hard. These labels describe the complexity of a specific instruction within a session, while the five benchmark difficulty levels describe the global scenario complexity. This distinction is important because realistic missions often contain both simple operational primitives and difficult collaborative subtasks. A robust LLM agent should be able to solve low-level operational instructions reliably while also adapting to harder planning requirements introduced by larger fleets, denser environments, heterogeneous target and obstacle layouts, and limited observability.

\begin{table}[htbp]
\centering
\caption{Overview of the benchmark. Each scenario family contains five difficulty levels (Easy, Intermediate, Moderate, Hard, and Extreme), with five sessions per level and 20 tasks per session.}
\label{tab:benchmark_overview}
\small
\setlength{\tabcolsep}{3pt}
\resizebox{\columnwidth}{!}{%
\begin{tabular}{lcccccc}
\hline
\textbf{Scenario Type} & \textbf{Sessions} & \textbf{Drones} & \textbf{Targets} & \textbf{Obstacles} & \textbf{Tasks} & \textbf{Checks} \\
\hline
Area Search & 25 & 3--20 & 3--15 & 6--100 & 500 & 3009 \\
Area Assignment and Patrol & 25 & 5--30 & 3--20 & 6--100 & 500 & 2813 \\
Target Assignment & 25 & 5--30 & 3--100 & 6--100 & 500 & 3574 \\
\hline
\textbf{Total} & \textbf{75} & 3--30 & 3--100 & 6--100 & \textbf{1500} & \textbf{9396} \\
\hline
\end{tabular}
}
\end{table}
 
\subsubsection{Evaluation Protocol and Metrics}
\label{sec:evaluation_protocol_metrics}

The benchmark evaluates each natural-language task with hidden platform-side checks. The primary metric is \textit{Task Pass Rate}, which counts a task as successful only when its validation logic passes. Because complex tasks may contain multiple checks, we also report check-level metrics to measure partial progress. These check-level metrics are important because long-horizon tasks may fail as full tasks while still completing part of the required validation chain.

\textit{Average Check Pass Rate} (Avg. Check) averages the internal check pass rate of each task, giving every task equal weight. \textit{Global Check Pass Rate} (Global Check) aggregates passed checks over all checks and is therefore weighted toward tasks with more validation conditions. \textit{Total Failed Task Rate} measures tasks for which no check passes, separating complete failures from partially completed tasks. Table~\ref{tab:benchmark_metrics} summarizes these metrics.

\begin{table}[htbp]
\centering
\caption{Evaluation metrics supported by the benchmark. Avg. Check is the per-task average check pass rate, while Global Check is the aggregate pass rate over all validation checks.}
\label{tab:benchmark_metrics}
\small
\setlength{\tabcolsep}{4pt}
\resizebox{\columnwidth}{!}{%
\begin{tabular}{lll}
\hline
\textbf{Metric} & \textbf{Level} & \textbf{Definition} \\
\hline
Task Pass Rate & Task & Passed tasks / total tasks \\
Avg. Check Pass Rate & Task & Mean of task-level check pass rates \\
Total Failed Rate & Task & Tasks with zero passed checks / total tasks \\
Global Check Pass Rate & Check & Passed checks / total checks across all tasks \\
\hline
\end{tabular}
}
\end{table}

Beyond aggregate scoring, the structure of the benchmark provides a diagnostic path for measuring progress in LLM-based multi-UAV planning. Simple tasks evaluate whether an agent can reliably translate language into basic UAV actions. Intermediate tasks test navigation, inspection, and sequential operation. Harder sessions introduce larger fleets, more targets, denser obstacles, area coverage, patrol, and collaborative assignment. This progression makes it possible to identify where current LLM agents succeed or fail, and to compare future methods under a shared task distribution and scoring protocol.

By combining natural-language task descriptions, hidden automatic validators, local perception constraints, and progressively harder collaborative scenarios, the benchmark establishes a new evaluation setting for LLM-oriented aerial robotics. It is closer to realistic UAV operation than full-information prompting because the agent must actively use UAVs as sensing and acting entities rather than assuming direct access to the complete world state. Therefore, the benchmark contributes a reusable foundation for future research on LLM-controlled multi-UAV autonomy, including agent workflow design, planning under partial observability, task allocation, and collaborative mission execution.

\subsection{Framework}

\begin{figure*}[!t]
\centering
\includegraphics[width=0.88\textwidth]{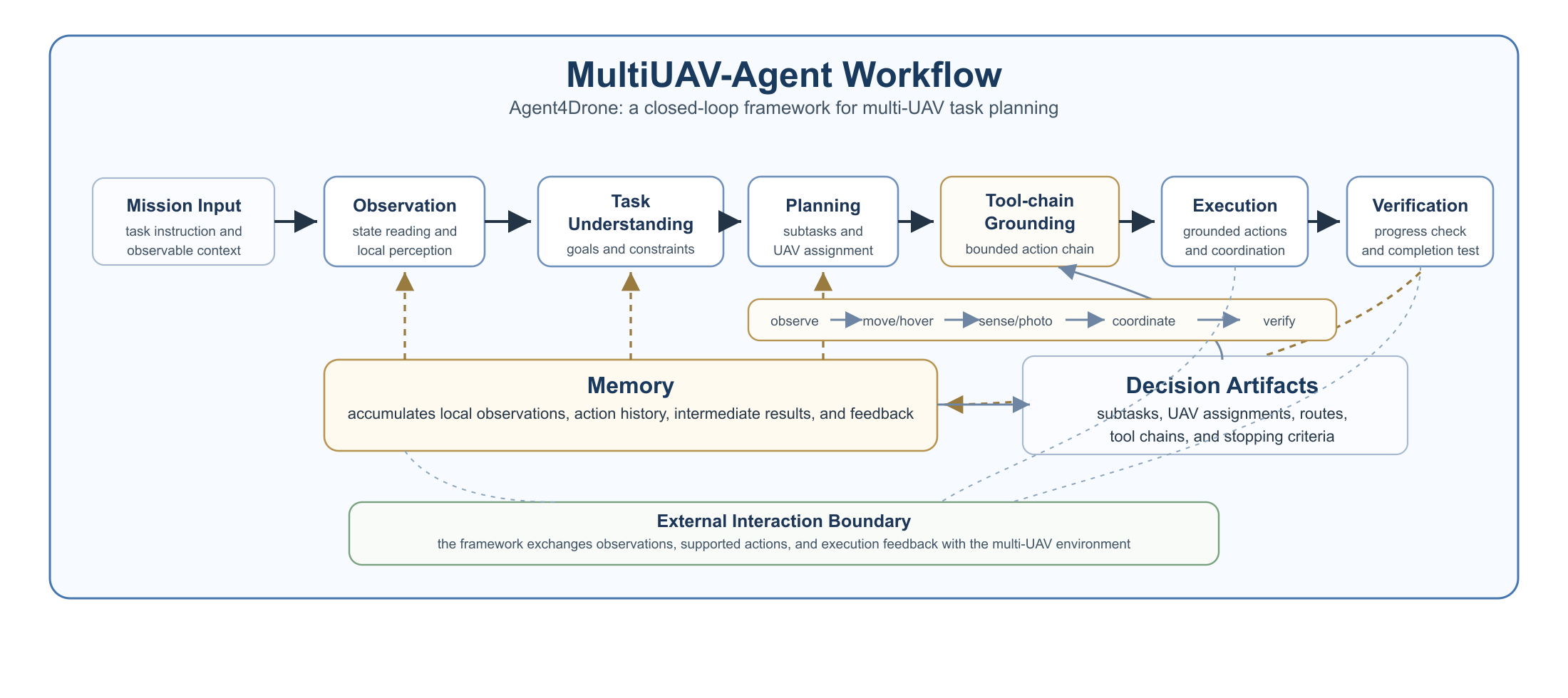}
\caption[Framework structure of Agent4Drone.]{Framework structure of Agent4Drone. The MultiUAV-Agent Workflow converts mission input and observable context into closed-loop observation, task understanding, planning, tool-chain grounding, execution, and verification. Memory stores local observations, action history, intermediate results, and feedback, while decision artifacts organize subtasks, UAV assignments, bounded tool/action chains, and stopping criteria for feedback-driven replanning.}
\label{fig:agent4drone_framework}
\end{figure*}

To provide an effective and reproducible framework for the benchmark, we propose a lightweight MultiUAV-Agent Workflow for LLM-based multi-UAV task planning. We refer to the proposed workflow as \textit{Agent4Drone}. The workflow is designed specifically for MultiUAV-Plat rather than as a generic prompting strategy. General-purpose LLM workflows such as chain-of-thought prompting or ReAct can improve reasoning and tool use, but they do not explicitly model key requirements of multi-UAV collaboration, including partial local perception, UAV selection, task decomposition, coordinated action execution, and state-based verification. In contrast, Agent4Drone organizes the agent around the platform's mission structure and restricted information boundary, making it easy to implement while still capturing the essential challenges of LLM-controlled aerial collaboration.

The workflow is intentionally positioned at the mission-planning layer rather than as a low-level UAV controller. It does not address flight stabilization, continuous trajectory optimization, or formal motion planning. Instead, it transforms a natural-language task and the currently visible scenario state into a short executable action sequence, executes that sequence through the platform, and uses feedback to decide whether to finish, retry, or replan. This design makes Agent4Drone suitable for systematic comparison, because it operates under the same platform, benchmark, visibility constraints, and validation protocol as generic agent baselines.

\subsubsection{Problem Formulation}

We formulate each benchmark task as a partially observable multi-UAV decision problem. A task is specified by a natural-language instruction $T$, an agent-access observation $O_t$ at step $t$, a platform-supported action set $\mathcal{A}$, and a hidden validation function $V$. The observation $O_t$ contains only information available to the agent, including task text, visible UAV states, command feedback, local perception, and accumulated interaction history. It excludes privileged scenario information, such as global target lists, global obstacle lists, and validation rules. The action set $\mathcal{A}$ contains constrained action families for observation, motion, sensing and photo capture, communication and coordination, and verification feedback.

An LLM-based multi-UAV agent can be described as a closed-loop decision rule:
\begin{equation}
\hspace*{\fill}
\pi_{\theta}: (T, O_t, M_t) \rightarrow (q_t,C_t),
\hspace*{\fill}
\end{equation}
where $M_t$ denotes the agent's memory, $q_t$ is the next high-level subtask, and $C_t$ is the corresponding bounded executable tool chain. The tool chain contains a finite sequence of platform-supported action steps:
\begin{equation}
\hspace*{\fill}
C_t=(c_t^1,\ldots,c_t^{L_t}),
\hspace*{\fill}
\end{equation}
where the superscript $l$ denotes the step index within the tool chain. Each step is defined as
\begin{equation}
\hspace*{\fill}
c_t^l=(u_t^l,a_t^l,p_t^l),
\hspace*{\fill}
\end{equation}
where $u_t^l$ is the responsible UAV or UAV group, $a_t^l \in \mathcal{A}$ is a platform-supported action, and $p_t^l$ contains the action parameters. The high-level subtask is grounded into its executable tool chain:
\begin{equation}
\hspace*{\fill}
q_t \Rightarrow C_t.
\hspace*{\fill}
\end{equation}
After executing a tool-chain segment, the agent receives execution and verification feedback $r_t$ and updates its observation and memory:
\begin{equation}
\hspace*{\fill}
(O_{t+1},M_{t+1})=\Phi(O_t,M_t,C_t,r_t).
\hspace*{\fill}
\end{equation}
The agent then applies the decision rule again with the updated observation and memory. Repeated application over $t=1,2,\ldots$ induces the full task-solving trajectory. The objective is to maximize task completion under the hidden validator $V$ while respecting platform-supported actions, agent-level information access, and bounded execution and replanning. This formulation differs from full-information planning because the agent may need to actively gather missing information through UAV movement and local perception before it can complete or verify the task.

\subsubsection{Workflow Design}

As shown in Fig.~\ref{fig:agent4drone_framework}, the Agent4Drone framework consists of six conceptual modules: Observation, Task Understanding, Memory, Planning, Execution, and Verification. These modules are not intended to be heavyweight software components; rather, they define a minimal structure that makes the LLM agent's reasoning process more stable, interpretable, and reproducible.

The Observation module gathers the current task description, UAV status, battery level, position, and local perception information. Since targets and obstacles are only visible within a UAV's perceived radius, observation is an active part of planning rather than a one-time query. 

The Task Understanding module converts each instruction into a structured interpretation. It identifies the likely task intent, candidate action type, visible task objects, spatial constraints, and whether the task requires one or multiple UAVs.

The Memory module uses a shared blackboard mechanism through which the framework modules read and update compact state across reasoning turns. The blackboard records the current task, visible environment context, UAV states and assignments, locally perceived entities, action and command outcomes, previous verification results, and retry status. It is restricted to information available through the agent-access interface and does not contain hidden validation rules or globally inaccessible targets and obstacles. This keeps the framework consistent with the benchmark's partial-observation setting while allowing observation, planning, execution, and verification to coordinate through a common state representation.

The Planning module selects appropriate UAVs and decomposes the task into a short sequence of subtasks with expected state changes. UAV selection considers visible position, battery level, operational status, perceived nearby entities, proximity to relevant visible targets, and whether the task benefits from collaboration. For simple operational tasks, the framework prefers a single primary UAV to avoid unnecessary coordination overhead. For area coverage, patrol, formation, or explicitly multi-UAV instructions, the framework assigns multiple UAVs while avoiding conflicting commands and redundant coverage. The plan is deliberately short-horizon: after meaningful actions, the agent re-observes the environment and updates memory before continuing.

The Execution module grounds each planned subtask into a bounded tool chain using supported action families, including observation, motion, sensing and photo capture, communication or coordination, and verification feedback. Agent4Drone executes a small number of actions, observes the updated state, verifies progress, and then decides whether to continue, retry, or replan. The Verification module compares expected and observed state changes and uses platform-side task validation feedback to decide whether the task has passed. If validation fails, the framework analyzes the discrepancy between expected and observed states, updates memory, and either retries or replans within a bounded budget.

\subsubsection{Closed-Loop Execution}

The complete framework operates as a closed loop. First, the agent observes the current task and visible session context. Second, it understands the task intent and identifies relevant constraints. Third, it selects suitable UAVs and generates a short subtask sequence. Fourth, it executes one or a small number of actions. Fifth, it re-observes the scenario and verifies whether the task has been satisfied. If the task passes, the framework records success and proceeds to the next task. If the task does not pass, the framework performs bounded retry or replanning based on updated observations.

This closed-loop design is important for MultiUAV-Plat because command success and task success are not equivalent. A command may be accepted by the platform while the task remains incomplete due to insufficient proximity, missing perception, incorrect UAV selection, incomplete area coverage, or failed coordination. By alternating short-horizon planning with observation and verification, the framework reduces the risk of long speculative action chains and allows the agent to recover from partial failures.

Failure handling follows conservative principles. If a command fails or verification remains negative, the agent first refreshes visible state rather than assuming the cause of failure. If repeated attempts do not improve the task state, the framework records a failure reason, such as missing target perception, insufficient movement, low battery, incorrect UAV assignment, or unmet collaboration condition. When uncertainty is high, the agent prefers safe fallback actions such as hovering, returning home, or landing, rather than continuing uncontrolled exploration. These rules make the framework practical for reproducible experiments and provide interpretable failure traces for later analysis.

Overall, Agent4Drone provides a simple yet task-specific framework for LLM-driven multi-UAV collaboration. It is easy to implement because it relies only on agent-access information and platform-supported actions, but it is more suitable than generic LLM workflows because it explicitly models partial perception, UAV selection, collaborative decomposition, feedback-driven execution, and bounded replanning. As a result, it establishes a practical starting point for future research on LLM agents in MultiUAV-Plat and the benchmark.

\section{Experiments}

The experiments evaluate Agent4Drone against a generic ReAct agent under the same MultiUAV-Plat Benchmark protocol. The analysis first compares overall task completion and partial validation success, then examines whether the performance gains are consistent across scenario families and difficulty levels. It further studies task categories and generated templates to identify remaining difficult cases, and uses validation-endpoint statistics to characterize the main residual failure modes.

\subsection{Experimental Setup}

All experiments use the MultiUAV-Plat Benchmark as the evaluation environment. The full benchmark contains 75 sessions, 1500 tasks, and 9396 automatic validation checks. The benchmark is organized into three scenario families, Target Assignment, Area Search, and Area Assignment and Patrol, and each family contains five difficulty levels. Unless otherwise stated, agents are evaluated under the platform's AGENT role. Therefore, the agent can observe task descriptions, UAV states, command feedback, and local perception, but it cannot access hidden validation specifications, global target lists, global obstacle lists, or scenario-authoring APIs.

Each task is evaluated by the platform-side task checker. A task is counted as passed only when its hidden check logic is satisfied. API return status is not scored as a separate success metric: an accepted or failed API call affects the result through the task conditions that remain satisfied or unsatisfied. In addition to task-level success, we report check-level success to capture partial progress inside complex tasks. For example, a multi-UAV formation task may fail as a whole while still satisfying several per-UAV movement or distance checks.

\textbf{Compared methods.} \textit{ReAct} is the baseline method used in the current full paired comparison. It follows a generic observe--reason--act loop where the model alternates between reading the visible task state, reasoning about the next step, issuing an action, and observing feedback. \textit{Agent4Drone} is our proposed task-specific framework. It augments closed-loop action with explicit memory, observation, task understanding, planning, execution, and verification modules designed for multi-UAV task allocation, local perception, and state-based recovery. Both methods run under the same AGENT role and use the same observation, action, and verification interfaces; neither method can access hidden validation specifications or privileged global state. To make the final evaluation more comprehensive, Table~\ref{tab:main_results} includes completed benchmark results under multiple backend LLMs.

\textbf{Metrics.} We use the task-level and check-level metrics defined in Section~\ref{sec:evaluation_protocol_metrics}, including Task Pass Rate, Average Task Check Pass Rate (Avg. Check), Global Check Pass Rate, and Total Failed Task Rate.
% Begin inlined from generated/overall_results.tex
\begin{table*}[htbp]
\centering
\caption{Main benchmark results for MultiUAV-Plat Benchmark across multiple backend LLMs. The Backend LLM column identifies the model used for each run. Values are generated by \texttt{tools/recompute\_results.py}.}
\label{tab:main_results}
\small
\setlength{\tabcolsep}{3pt}
\resizebox{\textwidth}{!}{%
\begin{tabular}{llcccccc}
\hline
\textbf{Method} & \textbf{Backend LLM} & \textbf{Passed Tasks} & \textbf{Total Failed Tasks} & \textbf{Task Pass} & \textbf{Avg. Check} & \textbf{Global Check} & \textbf{Total Failed} \\
\hline
ReAct & doubao-2-pro & 459 & 486 & 30.60\% & 47.91\% & 43.15\% & 32.40\% \\
ReAct & qwen3.5 & 629 & 333 & 41.93\% & 59.42\% & 56.29\% & 22.20\% \\
ReAct & deepseek-v4-pro & 954 & 118 & 63.60\% & 79.72\% & 73.09\% & 7.87\% \\
Agent4Drone & doubao-2-pro & 869 & 194 & 57.93\% & 74.58\% & 71.96\% & 12.93\% \\
Agent4Drone & deepseek-v4-flash & 1044 & 102 & 69.60\% & 83.40\% & 80.76\% & 6.80\% \\
Agent4Drone & deepseek-v4-pro & \textbf{1054} & \textbf{93} & \textbf{70.27\%} & \textbf{84.86\%} & \textbf{82.82\%} & \textbf{6.20\%} \\
\hline
\end{tabular}
}
\end{table*}
% End inlined from generated/overall_results.tex
\FloatBarrier

\subsection{Main Results}

Table~\ref{tab:main_results} summarizes the main completed benchmark results on the full MultiUAV-Plat Benchmark. In the paired doubao-2-pro comparison, Agent4Drone passes 869 of 1500 tasks, while ReAct passes 459 tasks. This improves task pass rate from 30.60\% to 57.93\%, a gain of 27.33 percentage points. Agent4Drone also improves Average Task Check Pass Rate from 47.91\% to 74.58\% and Global Check Pass Rate from 43.15\% to 71.96\%. With stronger DeepSeek backends, Agent4Drone reaches 70.27\% task pass using deepseek-v4-pro and 69.60\% using deepseek-v4-flash. The updated deepseek-v4-pro ReAct baseline reaches 63.60\%, higher than the doubao-2-pro and qwen3.5 ReAct baselines but still below Agent4Drone with the same backend.

For the deepseek-v4-pro full-benchmark run, the API usage records indicate an approximate total cost of CNY 400 (roughly USD 60) across more than 35K requests. The run used roughly 15M output tokens, 90M cache-miss input tokens, and 2B cache-hit input tokens. These figures provide a practical resource-consumption reference for executing the full 1500-task benchmark with Agent4Drone on deepseek-v4-pro.

The paired transition analysis in Fig.~\ref{fig:paired_transitions} compares the two backend LLMs for which both ReAct and Agent4Drone runs are available. Under doubao-2-pro, Agent4Drone converts 441 ReAct failures into passes while only 31 ReAct passes become failures. Under deepseek-v4-pro, the net gain is smaller but still positive: Agent4Drone converts 211 ReAct failures into passes, with 111 regressions.
% Begin inlined from generated/paired_transitions_figure.tex
\begin{center}
\centering
\includegraphics[width=\columnwidth]{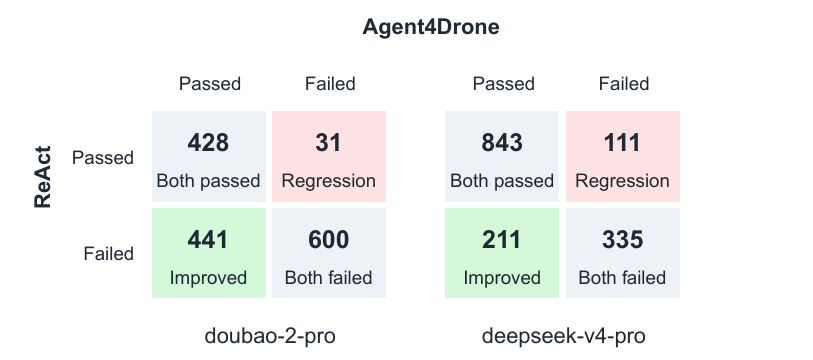}
\captionof{figure}{Paired task-level transition matrices between ReAct and Agent4Drone under doubao-2-pro and deepseek-v4-pro.}
\label{fig:paired_transitions}
\end{center}
% End inlined from generated/paired_transitions_figure.tex
\FloatBarrier

Figure~\ref{fig:scenario_difficulty_results} reports task pass rate by difficulty level across all completed method--backend runs. The curves show both model-capability effects and method effects: stronger backends improve the ReAct baseline substantially, while Agent4Drone remains competitive across all difficulty levels and is especially important on harder tasks for weaker backends.
% Begin inlined from generated/scenario_difficulty_figure.tex
\begin{center}
\centering
\includegraphics[width=\columnwidth]{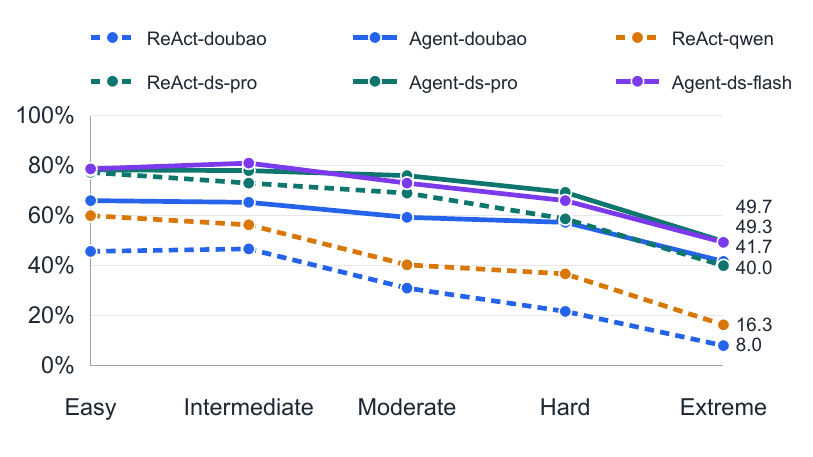}
\captionof{figure}{Task pass rate by difficulty level across all completed method--backend runs. Runs using the same backend LLM share a color; dashed lines denote ReAct and solid lines denote Agent4Drone.}
\label{fig:scenario_difficulty_results}
\end{center}
% End inlined from generated/scenario_difficulty_figure.tex
\FloatBarrier

Figure~\ref{fig:scenario_family_paired_changes} summarizes scenario-family task pass rates across all completed method--backend runs. Hue identifies the backend LLM, while darker cells indicate higher pass rates. Within the paired doubao-2-pro and deepseek-v4-pro columns, Agent4Drone improves over ReAct for every scenario family; the improvement is larger with doubao-2-pro, while deepseek-v4-pro still shows stable gains. Target Assignment remains the hardest family, indicating that stronger backend reasoning helps but does not remove target-pairing and target-verification difficulty.
% Begin inlined from generated/scenario_family_paired_changes_figure.tex
\begin{center}
\centering
\includegraphics[width=\columnwidth]{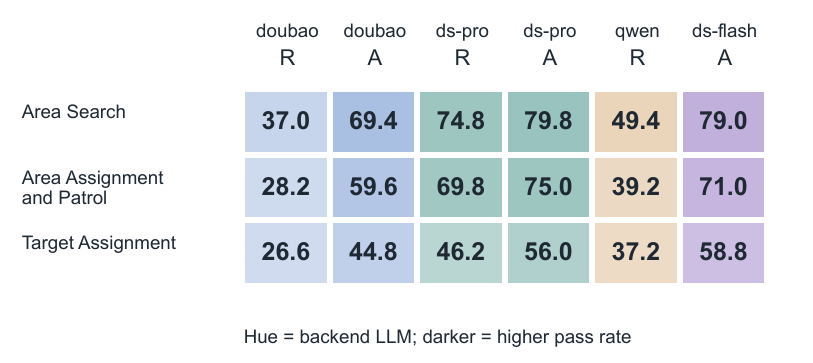}
\captionof{figure}{Scenario-family task pass rates across completed method--backend runs. Hue denotes the backend LLM and shade intensity denotes the pass rate.}
\label{fig:scenario_family_paired_changes}
\end{center}
% End inlined from generated/scenario_family_paired_changes_figure.tex
\FloatBarrier

\subsection{Task-Type and Template Analysis}

We analyze task types using the generator categories defined in the task-template system for the paired doubao-2-pro and deepseek-v4-pro runs. This is more stable than relying on surface task names alone because it follows the benchmark's intended task-generation structure. Table~\ref{tab:task_category_results} reports task pass rates for each method and backend. Agent4Drone is higher than ReAct in most categories under both backends, while Communication is already nearly saturated under deepseek-v4-pro.
% Begin inlined from generated/task_category_results.tex
\begin{table}[htbp]
\centering
\caption{Task pass rates by generated task category for paired backend runs.}
\label{tab:task_category_results}
\scriptsize
\setlength{\tabcolsep}{2.5pt}
\renewcommand{\arraystretch}{1.08}
\resizebox{\columnwidth}{!}{%
\begin{tabular}{lccccc}
\hline
\multirow{2}{*}{\textbf{Category}} & \multirow{2}{*}{\textbf{Tasks}} & \multicolumn{2}{c}{\fontsize{6}{7}\selectfont\textbf{doubao-2-pro}} & \multicolumn{2}{c}{\fontsize{6}{7}\selectfont\textbf{deepseek-v4-pro}} \\
\cline{3-6}
 & & \fontsize{6}{7}\selectfont\textbf{ReAct} & \fontsize{6}{7}\selectfont\textbf{Agent4Drone} & \fontsize{6}{7}\selectfont\textbf{ReAct} & \fontsize{6}{7}\selectfont\textbf{Agent4Drone} \\
\hline
Basic Operations & 146 & 79.45 & 93.84 & 86.99 & 97.95 \\
Navigation & 339 & 31.56 & 64.60 & 61.95 & 74.63 \\
Search & 396 & 23.74 & 53.79 & 69.70 & 70.20 \\
Multi-Drone Coordination & 514 & 17.12 & 39.88 & 48.05 & 54.09 \\
Delivery & 73 & 47.95 & 91.78 & 86.30 & 97.26 \\
Communication & 32 & 59.38 & 87.50 & 96.88 & 96.88 \\
\hline
\end{tabular}
}
\end{table}
% End inlined from generated/task_category_results.tex
Template-level results provide a finer view of these categories. Agent4Drone makes especially large gains on navigation-heavy and formation-heavy templates, including Fleet Coordinated Takeoff and Formation with five UAVs, Flight Marathon Navigation, Fly Several Waypoints, and Complex Obstacle Course Navigation. In contrast, target-assignment completion, area-search completion, sequential multi-target visits, and some patrol or sweep tasks remain difficult under stronger backends. Many of these hard templates show high average check completion but low strict task pass rate, indicating that Agent4Drone often completes part of the mission but misses at least one required UAV-target or fleet-level condition.

\subsection{Failure Analysis}

Endpoint-level validation for the paired runs reveals where the remaining failures occur. Table~\ref{tab:endpoint_failures} reports endpoint-level check pass rates for representative validation endpoints. The most stable improvements are movement and route-following checks: Agent4Drone reaches higher pass rates than ReAct on directed-distance and visited-position checks under both doubao-2-pro and deepseek-v4-pro. These improvements explain much of the gain on navigation, delivery, and formation templates.
% Begin inlined from generated/endpoint_failures.tex
\begin{table}[htbp]
\centering
\caption{Endpoint-level check pass rates for paired backend runs. Checks denotes the number of validation checks for each endpoint.}
\label{tab:endpoint_failures}
\scriptsize
\setlength{\tabcolsep}{2.5pt}
\renewcommand{\arraystretch}{1.08}
\resizebox{\columnwidth}{!}{%
\begin{tabular}{lccccc}
\hline
\multirow{2}{*}{\textbf{Endpoint}} & \multirow{2}{*}{\textbf{Checks}} & \multicolumn{2}{c}{\fontsize{6}{7}\selectfont\textbf{doubao-2-pro}} & \multicolumn{2}{c}{\fontsize{6}{7}\selectfont\textbf{deepseek-v4-pro}} \\
\cline{3-6}
 & & \fontsize{6}{7}\selectfont\textbf{ReAct} & \fontsize{6}{7}\selectfont\textbf{Agent4Drone} & \fontsize{6}{7}\selectfont\textbf{ReAct} & \fontsize{6}{7}\selectfont\textbf{Agent4Drone} \\
\hline
\makecell[l]{\texttt{drone\_has\_moved\_}\\\texttt{directed\_distance}} & 1917 & 48.83 & 90.14 & 77.99 & 98.85 \\
\makecell[l]{\texttt{drone\_has\_}\\\texttt{visited\_position}} & 3692 & 39.52 & 69.07 & 73.10 & 80.55 \\
\makecell[l]{\texttt{target\_is\_reached\_}\\\texttt{by\_drone}} & 1049 & 31.17 & 51.38 & 66.35 & 65.40 \\
\makecell[l]{\texttt{target\_in\_photo\_}\\\texttt{taken\_by\_drone}} & 465 & 14.62 & 37.20 & 40.00 & 51.61 \\
\texttt{drone\_position} & 214 & 23.83 & 56.07 & 49.07 & 69.63 \\
\texttt{task\_progress} & 83 & 21.69 & 10.84 & 77.11 & 50.60 \\
\texttt{target\_is\_reached} & 317 & 54.57 & 65.93 & 93.38 & 84.54 \\
\makecell[l]{\texttt{target\_is\_}\\\texttt{fully\_searched}} & 171 & 15.79 & 63.16 & 85.38 & 80.12 \\
\hline
\end{tabular}
}
\end{table}
% End inlined from generated/endpoint_failures.tex
The remaining bottlenecks are target grounding, final progress checks, and some target-completion endpoints. Under deepseek-v4-pro, Agent4Drone has lower pass rates than ReAct on \texttt{task\_progress}, \texttt{target\_is\_reached}, and \texttt{target\_is\_fully\_searched}, even though it improves movement, visited-position, photo, and drone-position checks. This suggests that future improvements should focus on active target localization, final-condition verification, and avoiding premature or inconsistent completion claims. The \texttt{task\_progress} comparison should be interpreted cautiously because this endpoint appears only 83 times.

\section{Conclusion}

This paper presented MultiUAV-Plat, an LLM-oriented simulation platform and benchmark for executable multi-UAV collaborative task planning. The platform exposes restricted UAV-control APIs, agent-level partial observations, local perception, optional visualization, and hidden task validators, enabling LLM agents to be evaluated through closed-loop interaction rather than privileged simulator state. On this basis, the MultiUAV-Plat Benchmark provides 75 mission sessions, 1500 tasks, and 9396 validation checks across target assignment, area search, and area assignment and patrol scenarios.

We also introduced Agent4Drone, a task-specific LLM-agent framework that structures multi-UAV behavior around observation, memory, planning, execution, and verification. In paired full-benchmark experiments, Agent4Drone improves task pass rate by 27.3 percentage points over ReAct under doubao-2-pro and remains competitive across stronger DeepSeek backends, with consistent check-level gains on movement and route-following tasks. The remaining failures highlight persistent challenges in target grounding, final-condition verification, and coordinated execution under partial observability. 

Future work will extend the platform and benchmark with richer dynamics, stronger target-localization strategies, ablation studies, and broader backend comparisons.


\bibliographystyle{cas-model2-names}


\bibliography{cas-refs}

% Appendix omitted from the current version.
% \clearpage
% \onecolumn
% \appendix
% \section*{Appendix}
% \addcontentsline{toc}{section}{Appendix}
% \section{Task-Name Family Results}
%
% Table~\ref{tab:appendix_task_name_results} reports performance for combined task-name families in the default template-file order for the paired doubao-2-pro runs. Numbered variants of the same generated task family are aggregated into one row.
%
% \input{generated/appendix_task_name_results.tex}
%
% \section{Task-Family Gains and Regressions}
%
% Table~\ref{tab:appendix_task_family_gains_regressions} reports the largest task-family gains and true regressions for the paired doubao-2-pro and deepseek-v4-pro runs.
%
% \input{generated/appendix_task_family_gains_regressions.tex}


% Author biography without author photo.


% \bio{figs/cas-pic1}
% Author biography with author photo.


% \bio{figs/cas-pic1}
% Author biography with author photo.

\end{document}